\newif\ifuseiclr
\useiclrtrue          

\ifuseiclr
  \documentclass{article}
  \usepackage{iclr2026_conference,times}
\else
  \documentclass[letterpaper,10pt]{article}
  \usepackage{times}
  \usepackage[left=1.15in, right=1.15in, top=1.2in, bottom=1in, headsep=.25in]{geometry}
  \usepackage{fancyhdr}
  \usepackage{natbib}
  \setcitestyle{authoryear,round,citesep={;},aysep={,},yysep={;}}
  \renewenvironment{abstract}
    {\small\begin{quote}\noindent\par{\sc \abstractname.}}
    {\noindent\end{quote}}
\fi

\usepackage{amsmath, amsthm, amssymb, mathtools}
\ifuseiclr\else\allowdisplaybreaks\fi
\usepackage{graphicx, caption, subcaption}
\usepackage{booktabs, multirow}
\usepackage{xcolor}
\usepackage{hyperref}
\definecolor{darkblue}{RGB}{0,51,153}
\hypersetup{
  colorlinks=true,
  linkcolor=darkblue,
  citecolor=darkblue,
  urlcolor=darkblue,
}
\usepackage{url}
\usepackage{microtype}
\usepackage{cleveref}
\usepackage{bm}
\usepackage{bbm}

\theoremstyle{definition}

\theoremstyle{remark}

\newcommand{\R}{\mathbb{R}}

\ifuseiclr
  \iclrfinalcopy 
\fi

\title{\textbf{Universal Diffusion-Based Probabilistic Downscaling}}

\ifuseiclr
  \author{\mdseries
    \textbf{Roberto Molinaro}\textsuperscript{1}\thanks{Equal contribution. Correspondence to \href{mailto:firstname.lastname@jua.ai}{\emph{firstname}.\emph{lastname}@jua.ai}.}\quad
    \textbf{Niall Siegenheim}\textsuperscript{1}\footnotemark[1]\quad
    \textbf{Henry Martin}\textsuperscript{1}\quad
    \textbf{Mark Frey}\textsuperscript{1} \\[0.3em]
    \textbf{Niels Poulsen}\textsuperscript{1}\quad
    \textbf{Philipp Seitz}\textsuperscript{1}\quad
    \textbf{Marvin Vincent Gabler}\textsuperscript{1} \\[0.8em]
    \textsuperscript{1}Jua.ai
  }
\else
  \author{
    Roberto Molinaro\textsuperscript{1}\thanks{Equal contribution. Correspondence to \href{mailto:firstname.lastname@jua.ai}{firstname.lastname@jua.ai}.} \and
    Niall Siegenheim\textsuperscript{1}\footnotemark[1] \and
    Henry Martin\textsuperscript{1} \and
    Mark Frey\textsuperscript{1} \and
    Niels Poulsen\textsuperscript{1} \and
    Philipp Seitz\textsuperscript{1} \and
    Marvin Vincent Gabler\textsuperscript{1}
  }
  \date{\textsuperscript{1}Jua.ai}
\fi

\begin{document}
\maketitle

\begin{abstract}
We introduce a universal diffusion-based downscaling framework that lifts
deterministic low-resolution weather forecasts into probabilistic
high-resolution predictions without any model-specific fine-tuning.
A single conditional diffusion model is trained on paired coarse-resolution
inputs ($\sim$25\,km resolution) and high-resolution regional reanalysis targets ($\sim$5\,km resolution),
and is applied in a fully zero-shot manner to deterministic forecasts from
heterogeneous upstream weather models. Focusing on near-surface variables, we evaluate probabilistic forecasts against
independent in situ station observations over lead times up to 90\,h.
Across a diverse set of AI-based and numerical weather prediction (NWP) systems,
the ensemble mean of the downscaled forecasts consistently improves upon each
model's own raw deterministic forecast, and substantially larger gains are
observed in probabilistic skill as measured by CRPS.
These results demonstrate that diffusion-based downscaling provides a scalable,
model-agnostic probabilistic interface for enhancing spatial resolution and
uncertainty representation in operational weather forecasting pipelines.
\end{abstract}

\section{Introduction}
Accurate weather prediction requires approximating solutions of highly nonlinear
partial differential equations governing atmospheric flow.
Operational global forecasting systems represent the atmosphere on finite grids,
typically at horizontal resolutions of 9--25\,km for global deterministic
forecasts.
At these resolutions, many near-surface processes and land--atmosphere
interactions remain only partially resolved.
Sub-grid-scale effects associated with terrain, surface roughness, land use, and
boundary-layer structure are therefore suppressed or parameterized, leading to
systematic loss of small-scale spatial variability.
This loss directly impacts forecast skill for near-surface variables such as
2\,m temperature and 10\,m wind speed, which are strongly controlled by local
surface heterogeneity \citep{maraun2010,benestad2008}.

A widely used strategy to address this limitation is \emph{downscaling}, i.e.\
the enrichment of coarse-resolution forecasts with fine-scale spatial detail.
Dynamical downscaling resolves the governing equations on a finer grid using a
nested regional model driven by coarse boundary conditions
\citep{giorgi1999}.
While physically consistent, this approach remains computationally expensive and
tightly coupled to specific NWP model configurations.

An alternative class of methods, commonly referred to as \emph{statistical} or
data-driven downscaling, seeks to infer fine-scale fields directly from
coarse-resolution predictors using empirical or learned mappings
\citep{wilby1998,maraun2010}.
From a modeling perspective, statistical downscaling constitutes an ill-posed inverse
problem: the mapping from a coarse atmospheric state to fine-scale realizations
is not unique, and multiple high-resolution fields may be consistent with the
same large-scale conditions.
As a result, deterministic regression-based approaches whether based on linear
models, analog methods, or supervised machine learning tend to converge toward
conditional means, producing overly smooth fields and systematically
under-representing variability and uncertainty
\citep{benestad2008,maraun2010}.
This limitation is particularly pronounced for near-surface variables and in
complex terrain.

In addition, access to probabilistic forecasts is essential for many weather-dependent decision-making applications. Nevertheless, operational systems are often restricted to deterministic outputs because uncertainty estimation, typically achieved via ensemble models is computationally expensive.

A natural remedy is to abandon deterministic mappings altogether and instead
formulate downscaling as the problem of learning the \emph{conditional
distribution} of fine-scale states given a coarse-resolution forecast.
However, learning high-dimensional conditional distributions in a stable and
scalable manner has remained a major challenge for data-driven downscaling.

Recent advances in generative modeling offer a promising pathway to address this
challenge.
In particular, diffusion-based models have demonstrated the ability to learn
complex, high-dimensional probability distributions and to generate realistic
samples conditioned on auxiliary inputs
\citep{ho2020ddpm,karras2022elucidating,molinaro2025generativeaifastaccurate}.
In this work, we adopt this perspective and formulate high-resolution
downscaling as a conditional generative modeling problem.
We introduce a diffusion-based framework that learns a stochastic mapping from
coarse-resolution atmospheric states to distributions of fine-scale surface
fields, trained solely on paired reanalysis data.

A key feature of our approach is its \emph{model-agnostic} design.
The downscaler is trained once using low-resolution$\rightarrow$high-resolution \textit{analysis or reanalysis} pairs and
is subsequently applied \emph{zero-shot} as a post-processing step to a diverse
set of \textit{heterogeneous deterministic forecasts}, spanning both AI-based and NWP
systems.
This decouples the learning of fine-scale surface structure from the upstream
forecast model and allows a single probabilistic module to be reused across
forecasting pipelines with different numerical formulations, resolutions, and
systematic biases.

Using independent in situ station observations, we show that diffusion-based
downscaling consistently improves forecast quality across all tested upstream
models.
Improvements are observed both in point accuracy, measured by RMSE of the
ensemble mean, and—more prominently—in probabilistic skill, measured by CRPS.
This demonstrates that diffusion-based downscaling enables uncertainty-aware probabilistic forecasting from deterministic inputs, rather than improving point predictions alone.

\paragraph{Our main contributions are:}
\begin{itemize}
    \item We introduce a \emph{single}, model-agnostic conditional diffusion
    downscaler trained solely on low-resolution$\rightarrow$high-resolution reanalysis pairs, which can
    be applied \emph{zero-shot} as a post-processing step to heterogeneous
    \textit{operational} deterministic forecasts from both AI-based and NWP models.
     \item We show that a single deterministic forecast can be lifted to a probabilistic forecast without explicit ensembling, providing uncertainty-aware predictions at a fraction of the computational cost of traditional ensemble systems.
    \item We demonstrate through independent weather station verification that this
    plug-and-play downscaling consistently improves forecast quality across
    diverse upstream systems \emph{at continental (European) scale}, yielding gains
    in both point accuracy and probabilistic skill across multiple near-surface
    variables.
\end{itemize}

\section{Related Work}

In recent years, there has been substantial interest in learning solution operators to partial differential equations from data, rather than solving the governing equations using numerical integration \citep{li_fourier_2021,raonic_convolutional_2023,herde_poseidon_2024-1}.
Weather and climate modeling has proven itself as a particularly fertile ground for such approaches, given the vast amounts of historical data available from reanalysis products \citep{hersbach_era5_2020} and the high computational cost of traditional numerical solvers \citep{pathak_fourcastnet_2022,lam_learning_2023,lang_aifs_2024,bodnar_foundation_2024}.

While many AI-based weather models have shown impressive results in terms of error metrics compared to their numerical counterparts such as the ECMWF IFS \citep{ecmwf_ifs_upgrade_181}, their deterministic training objectives (e.g., mean squared error) lead them to regress to the conditional mean of the target distribution, resulting in smoothed predictions that lack physically relevant small-scale structure \citep{selz_effective_2025}.
To address this limitation, recent works have explored using diffusion models to learn probabilistic weather forecasts that capture the full distribution of possible outcomes \citep{price_probabilistic_2025,cachay_elucidated_2025}. However, these large models are computationally expensive during inference time.

In practice, short-term high-resolution regional weather forecasts are often preferred to global ones, as they can better resolve local features such as orography, land-sea contrasts, and surface heterogeneity that strongly influence near-surface variables. The most notable examples are NOAA's High Resolution Rapid Refresh (HRRR) model \citep{dowell_high-resolution_2022} and ICON-EU, a European model operated by DWD \citep{zangl_icon_2015}.
These operate as limited-area models (LAMs), which numerically integrate the governing equations over a restricted spatial domain using boundary conditions provided by a global model.

In the case of HRRR, there have been attempts at using machine learning techniques to obtain similar improvements as in global weather forecasting models by replacing the numerical time integration with a learned diffusion-based model \citep{pathak_kilometer-scale_2024,abdi_hrrrcast_2025}. However, they continue to rely on a global numerical weather prediction model for their boundary conditions.
While there has been research on more localised AI-based LAMs for Europe \citep{oskarsson_graph-based_2023,adamov_building_2025-2}, to the best of our knowledge, there have been no published attempts at building an AI-based LAM which covers the whole European domain as ICON-EU does.

Concurrently, statistical downscaling techniques focus on enhancing the resolution of coarse global weather states by learning a mapping from low-resolution inputs to high-resolution outputs using historical data, without incorporating any time integration. Dating back to the last century \citep{wilby_downscaling_1997,wilby_statistical_1998}, machine learning-based approaches have accelerated this line of research in recent years \citep{leinonen_stochastic_2021,mardani_residual_2024,wan_debias_2023,bischoff_unpaired_2024,merizzi_wind_2024,wan_regional_2025,tomasi_can_2025-1,glawion_global_2025,schillinger_enscale_2025}. Most of these works use diffusion models to learn these mappings, given their success in generating realistic images in computer vision \citep{rombach_high-resolution_2022}.

While these efforts are formidable, we believe these works have yet to fully explore the potential of diffusion-based downscaling as a universal, model-agnostic interface between upstream forecasting systems and high-resolution probabilistic predictions. Concretely, they suffer from one or more of the following limitations:
(i) They do not evaluate their downscaling models when used for operational high-resolution forecasting.
(ii) They focus on smaller regions (e.g. Italy \citep{merizzi_wind_2024,tomasi_can_2025-1}, Taiwan \citep{mardani_residual_2024}) rather than continental scales.
(iii) They focus on downscaling single variables (e.g. precipitation \citep{leinonen_stochastic_2021,glawion_global_2025}).

In this work, we address all of these limitations by exploring statistical downscaling through diffusion models as a means of \emph{improving operational medium-range weather forecasts} on a continental scale across multiple relevant surface variables.


\section{Problem Formulation}
\label{sec:problem}

The large-scale atmospheric state evolves according to the hydrostatic primitive
equations, which describe the conservation of momentum, mass, and energy.
Following \citet{holton_introduction_2012}, let $\mathbf{v}$ denote the horizontal velocity, $T$ the temperature, $p$ the
pressure, and $\rho$ the density.
Using pressure as the vertical coordinate and writing $\nabla$ for the horizontal
gradient, the governing equations are
{\allowdisplaybreaks
\begin{align}
    \label{eq:momentum}
    \frac{\partial \mathbf{v}}{\partial t}
    + (\mathbf{v} \cdot \nabla)\mathbf{v}
    + f\,\mathbf{k} \times \mathbf{v}
    &= -\frac{1}{\rho}\nabla p + \mathbf{F}, \\
    \label{eq:continuity}
    \frac{\partial \rho}{\partial t}
    + \nabla \cdot (\rho \mathbf{v})
    &= 0, \\
    \label{eq:thermo}
    \frac{\partial T}{\partial t}
    + \mathbf{v} \cdot \nabla T
    &= \frac{Q}{c_p}
    + \frac{R T\,\omega}{c_p p}, \\
    \label{eq:hydrostatic}
    \frac{\partial p}{\partial z}
    &= -\rho g, \qquad
    p = \rho R T ,
\end{align}}
where $\omega \equiv Dp/Dt$ denotes the pressure vertical velocity.
In height coordinates, the material derivative reads
\begin{equation}
\label{eq:material_derivative_z}
\frac{D}{Dt}
=
\frac{\partial}{\partial t}
+
\mathbf{v}\cdot\nabla
+
w\,\frac{\partial}{\partial z},
\end{equation}
yielding
\begin{equation}
\label{eq:omega_from_w}
\omega
=
\frac{\partial p}{\partial t}
+
\mathbf{v}\cdot\nabla p
-
\rho g\, w ,
\end{equation}
after invoking hydrostatic balance.

In practice, operational forecasting systems approximate solutions of
\cref{eq:momentum,eq:continuity,eq:thermo,eq:hydrostatic}
on discrete grids.
Let $\bar{u} \in \mathcal{U}_{\mathrm{LR}}$ denote a coarse-resolution atmospheric
state produced by an upstream forecasting system, and let
$u \in \mathcal{U}_{\mathrm{HR}}$ denote a corresponding high-resolution surface
state defined on a finer grid. The downscaling problem consists in inferring plausible realizations of $u$ given
$\bar{u}$.
Since the mapping $\bar{u} \mapsto u$ is not injective, this inference problem is
intrinsically non-unique.
We therefore formulate downscaling as a conditional density estimation task and
seek to learn the distribution $p(u \mid \bar{u})$, rather than a single deterministic mapping.

Learning this conditional distribution relies on training data consisting of
paired low- and high-resolution analysis or reanalysis fields
$\mathcal{D} = \{(\bar{u}_i, u_i)\}_{i=1}^N$ that are aligned in time and represent the same
underlying atmospheric state at different spatial resolutions.
Each pair constitutes a \emph{single} high-resolution realization associated
with its coarse counterpart; multiple realizations $u$ for the same
$\bar{u}$ are not observed.
The conditional distribution is therefore not available empirically but must be
inferred statistically from the ensemble of such pairs across time.

\begin{figure}[t]

    \centering
    \includegraphics[width=0.95\linewidth]{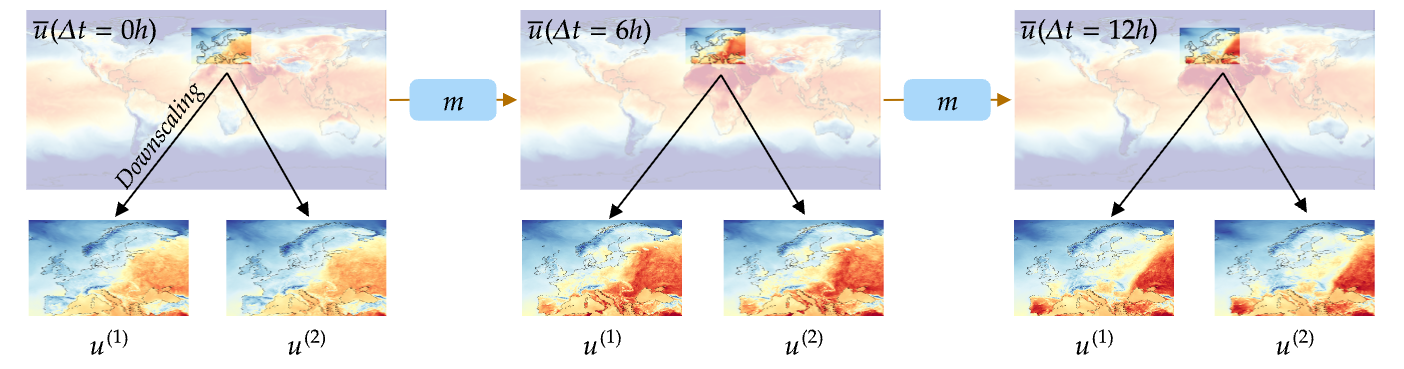}
\caption{
Model-agnostic conditional diffusion downscaling.
A coarse-resolution forecast $\bar{u}$, produced by different upstream models
$m \in \mathcal{M}$, conditions a reverse diffusion process.
Integrating the probability-flow dynamics yields multiple high-resolution
realizations sampled from $p_\theta(u \mid \bar{u})$.
}
\label{fig:universal_diffuser}
\end{figure}

At inference time, the conditioning input $\bar{u}$ may originate from \textit{any
deterministic forecasting system} and from \textit{any forecast lead
time} $\Delta t$.
Let $m \in \mathcal{M}$ index such a system, producing a coarse forecast
$\bar{u}^{(m)}$ on its native grid and variable conventions.
Our objective is to learn a single conditional model
$p_\theta(u \mid \tilde{u})$ that can be applied zero-shot to all
$m \in \mathcal{M}$ (see Figure \ref{fig:universal_diffuser} for a schematic representation of the process).
This formulation yields a probabilistic downscaler that produces ensembles of high-resolution realizations.
The ensemble mean provides a point forecast, while the ensemble variability captures uncertainty arising from unresolved small-scale structure.

\begin{figure}[t]
    \centering
    \includegraphics[width=\linewidth]{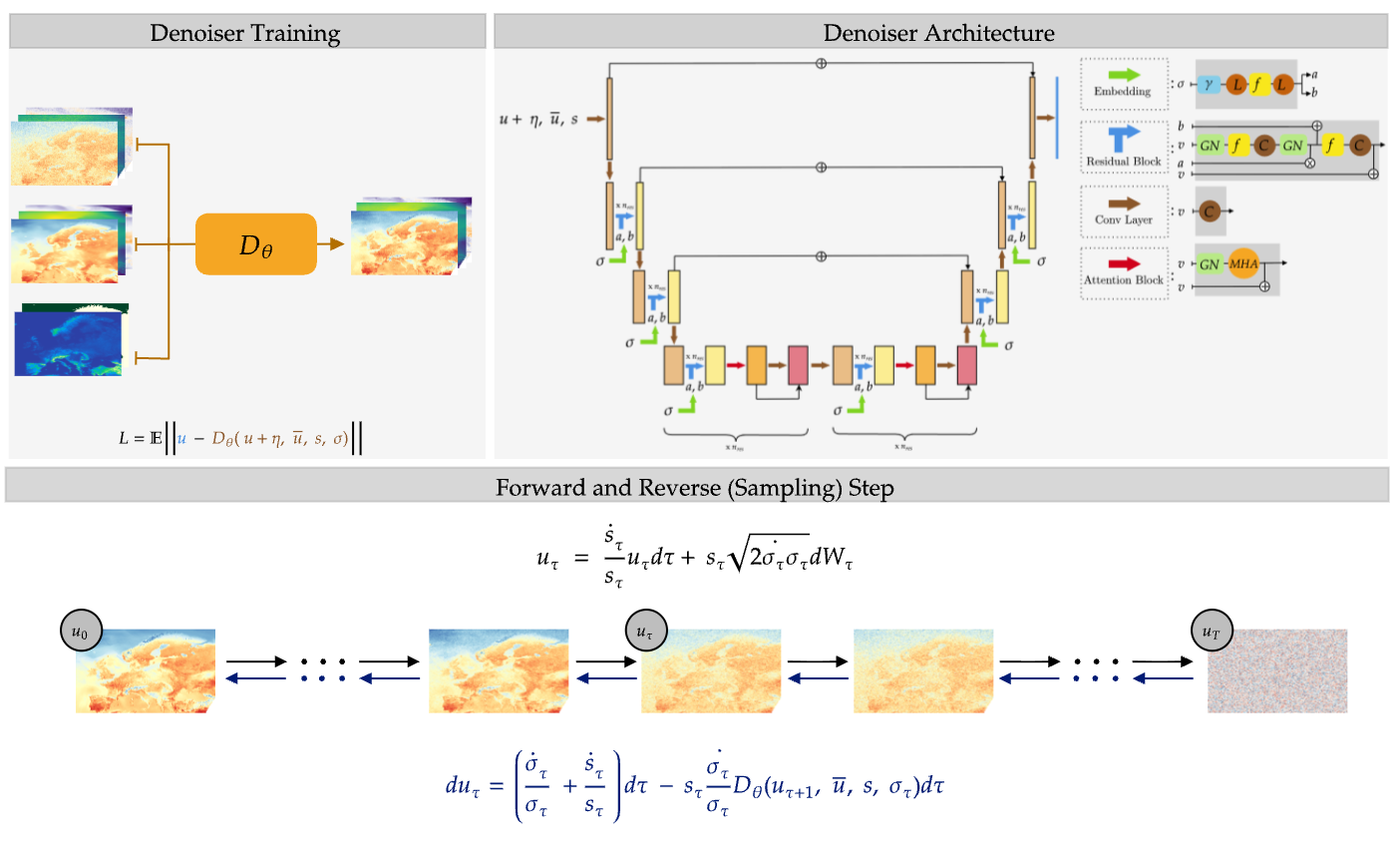}
    \caption{\textbf{Diffusion-based downscaling architecture and sampling procedure.}
    \emph{Top left:} Denoiser training.
    A clean high-resolution target $u$ is corrupted with Gaussian noise
    $\eta \sim \mathcal{N}(0,\sigma_\tau^2 I)$ and, together with the conditioning
    inputs, passed to the denoiser $D_\theta$, which is trained to reconstruct $u$
    via a mean-squared denoising objective.
    \emph{Top right:} Denoiser architecture.
    A U-Net encoder--decoder with residual blocks at each resolution and attention at
    the bottleneck.
    The noisy sample $u_\tau$ is concatenated with the coarse-resolution forecast
    $\bar{u}$ and static high-resolution
    fields.
    The noise level $\sigma_\tau$ (and lead time, when applicable) is embedded and
    injected at all resolutions via adaptive scale--shift modulation.
    \emph{Bottom:} Forward and reverse diffusion processes.
    The forward process progressively corrupts $u$ into pure noise, while sampling
    integrates the reverse-time probability-flow dynamics to generate high-resolution
    samples from $p_\theta(u\mid\bar{u})$.}
    
    \label{fig:denoiser_schematic}
  \end{figure}

\section{Diffusion-Based Downscaling Model}

To address the learning task depicted in Section \ref{sec:problem}, we adopt a conditional generative modeling approach and represent
the distribution $p(u \mid \bar{u})$ using a score-based diffusion formulation, following \citet{song2021scorebased}. Instead of defining sampling from an explicit parametric density, we use a generative stochastic process.

Concretely, we embed the conditional data distribution into a family of
noise-perturbed conditionals indexed by an artificial diffusion time
$\tau \in [0,K]$.
Given paired samples $(\bar{u},u) \in \mathcal{D}$, the forward diffusion process
progressively corrupts the high-resolution target by additive Gaussian noise
according to $u_\tau = u + \eta, ~ \eta \sim \mathcal{N}(0,\sigma_\tau^2 I)$,
where $\sigma_\tau$ denotes a prescribed noise schedule.
For sufficiently large $\sigma_\tau$, the conditional distribution
$p_\tau(u_\tau \mid \bar{u})$ becomes indistinguishable from an isotropic Gaussian,
providing a tractable terminal distribution from which sampling can be initiated.

The generative task then consists of reversing this noising process, transforming
a sample $u_K$ drawn from the terminal noise distribution into a sample from the
target conditional distribution $p(u \mid \bar{u})$.
In the continuous-time limit, this reverse dynamics is described by the
(backward-time) stochastic differential equation
\begin{equation}
\label{eq:reverse_sde_downscaling}
\mathrm{d}u_\tau
=
-\,2\,\dot{\sigma}_\tau \sigma_\tau
\nabla_{u_\tau} \log p_\tau(u_\tau \mid \bar{u})\,\mathrm{d}\tau
+
\sqrt{2\,\dot{\sigma}_\tau \sigma_\tau}\,\mathrm{d}\widehat{W}_\tau,
\end{equation}
where $\widehat{W}_\tau$ denotes Brownian motion in reverse diffusion time.

While the reverse SDE is known, it is not directly tractable in practice because
it depends on the conditional score
$\nabla_{u_\tau} \log p_\tau(u_\tau \mid \bar{u})$, which is not available in
closed form.
Following standard score-based diffusion models, we approximate this score using
a noise-conditional denoising operator $D_\theta(u_\tau,\bar{u},\sigma_\tau)$, implemented as a neural network with trainable parameters~$\theta$.
In particular, the score is approximated using Tweedie's formula as
\begin{equation}
\label{eq:tweedie1}
    \nabla_{u_\tau} \log p_\tau(u_\tau \mid \bar{u})
    \approx
    \frac{D_\theta(u_\tau,\bar{u},\sigma_\tau) - u_\tau}{\sigma_\tau^2}.
\end{equation}

Accordingly, the denoiser is trained to predict the clean high-resolution state
from its noisy counterpart by minimizing a mean-squared denoising objective over
paired data and over a continuous range of noise levels,
\begin{equation}
    \label{eq:loss1_downscaling}
    \mathcal{L}(D_\theta)
    =
    \mathbb{E}_{\sigma \sim p_{\mathrm{train}}}
    \mathbb{E}_{(\bar{u},u)\sim \mathcal{D}}
    \mathbb{E}_{\eta \sim \mathcal{N}(0,\sigma^2 I)}
    \bigl\|
        D_\theta(u+\eta,\bar{u},\sigma) - u
    \bigr\|_2^2.
\end{equation}
In practice, we follow the preconditioning and implementation choices introduced
in the Elucidated Diffusion Models framework \citep{karras2022elucidating}; full architectural and training
details are provided in Appendix~\ref{sec:appendix}.

\section{Experimental Setup and Results}
\label{sec:experiments}

This section describes the experimental setup used to evaluate the proposed
diffusion-based downscaling framework.
We detail the training data, model configuration, and evaluation protocol, and
assess probabilistic high-resolution forecasts against independent in situ
station observations across multiple heterogeneous upstream forecasting models.

\textbf{Training data.}
For training, we choose ERA5 reanalysis \citep{hersbach_era5_2020} fields as low-resolution inputs $\bar{u}$ ($\sim$25\,km / 0.25$^\circ$ resolution)
and CERRA reanalysis \citep{ridal_cerra_2024} fields as the corresponding high-resolution targets $u$ ($\sim$5\,km / 0.05$^\circ$ resolution).
ERA5 is well-known global atmospheric reanalysis produced by the European Centre for
Medium-Range Weather Forecasts (ECMWF), combining a state-of-the-art numerical
weather prediction model with a comprehensive set of in situ and satellite
observations.
CERRA is a high-resolution regional reanalysis for Europe and provides surface fields at approximately 5\,km horizontal resolution,
covering latitudes from 36$^\circ$ to 72$^\circ$\,N and longitudes from
15$^\circ$\,W to 45$^\circ$\,E.
Both ERA5 inputs and CERRA targets include 2\,m temperature (t2m), eastward and
northward 10\,m wind components (u10, v10), and mean sea-level pressure (msl).
Training is performed on data spanning the period 2014--2023.
All input and output channels are standardized using ERA5 
statistics computed over the period 1979--2021.

\textbf{Denoiser training, architecture, and conditioning interface.}
Figure~\ref{fig:denoiser_schematic} provides an overview of the denoiser training,
architecture, and sampling procedure.

The denoiser $D_\theta$ is implemented as a U-Net operating on 2D
latitude--longitude fields, augmented with attention at the bottleneck to enable
global spatial mixing.
The denoiser takes as input the noisy
high-resolution sample $u_\tau$ on the CERRA grid, the corresponding
coarse-resolution ERA5 data $\bar{u}$, and static high-resolution fields
$s=(z,\mathrm{lsm})$ consisting of elevation $z$ and land--sea mask
$\mathrm{lsm}$.
During training, $\bar{u}$ corresponds to ERA5, while at inference time it is
provided by a forecasting model.

To account for systematic differences in effective resolution, spectral content,
and implicit smoothing across upstream models (particularly over
long forecast horizons, where autoregressive AI models are known to progressively
smooth their predictions) we augment the conditioning input during training by
applying randomized spectral low-pass filtering to the ERA5 fields
(see Appendix~\ref{sec:appendix:smoothing} for details).

The coarse-resolution atmospheric state is bilinearly upsampled to the CERRA grid and
concatenated channel-wise with the noisy sample and static fields, yielding the
network input
\[
x_\tau
=
\mathrm{Concat}\!\bigl[
u_\tau,\;
\bar{u},\;
s
\bigr]
\in \mathbb{R}^{H \times W \times C}.
\]
where $H=721$, $W=1201$ and $C=10$.
The diffusion noise level $\sigma_\tau$ is embedded using Fourier features and
processed by a small multilayer perceptron to produce per-channel scale and shift
parameters.
These parameters modulate normalization layers within each residual block in a
FiLM-style manner and are applied at all resolutions throughout the encoder and
decoder (see Appendix~\ref{sec:appendix} for further details on training, inference and denoiser architecture).

\textbf{Evaluation protocol.}
The model is evaluated in a strictly out-of-sample setting spanning the period
from 1~July~2024 to 30~June~2025.
Forecast skill is assessed against roughly 10.000 independent in situ observations, focusing on
2\,m temperature and 10\,m wind speed.
We use a curated subset of high-quality station networks, including WMO surface
synoptic observations (SYNOP), WIS~2.0 surface observations, and aviation METAR
reports (including ASOS/AWOS and international METAR feeds).
Gridded forecasts are collocated to station locations using bilinear
interpolation in latitude and longitude.

Unless stated otherwise, probabilistic forecasts are produced using ensembles of
$n{=}16$ high-resolution samples.
Point forecast skill is quantified using the root-mean-square error (RMSE)
between observations $y$ and the ensemble mean
$\hat{y} = \frac{1}{n}\sum_{k=1}^n y^{(k)}$.
Probabilistic forecast quality is quantified using the continuous ranked
probability score (CRPS).
In practice, for an ensemble forecast
$\{y^{(k)}\}_{k=1}^n$ of size $n$, CRPS can be computed in closed form as
\begin{equation}
\label{eq:crps_ensemble}
\mathrm{CRPS}
=
\frac{1}{n} \sum_{k=1}^n \lvert y^{(k)} - y \rvert
-
\frac{1}{2n^2} \sum_{k=1}^n \sum_{\ell=1}^n
\lvert y^{(k)} - y^{(\ell)} \rvert .
\end{equation}
The first term measures the average absolute error of the ensemble members with respect to the observation, while the second term accounts for the internal dispersion of the ensemble. For a deterministic forecast $\hat{y}$ CRPS reduces exactly to the absolute error (and thus averages to MAE over
a dataset). 

\textbf{Upstream forecast models.}
We evaluate the proposed downscaling approach across a heterogeneous
set of upstream forecasting systems spanning both data-driven AI models and
physics-based numerical weather prediction (NWP) models:
Microsoft Aurora~\citep{bodnar_foundation_2024}, ECMWF AIFS~\citep{lang_aifs_2024},
Jua EPT-2~\citep{ept2_techreport}, NOAA GFS, and ECMWF IFS HRES.
Aurora, AIFS, and EPT-2 are learned forecasting systems trained on historical reanalysis and analysis data, while GFS and IFS HRES are operational NWP models based on numerical integration of the governing equations. 
Since Aurora and AIFS produce forecasts at a 6-hourly temporal resolution only, all models are evaluated at a common 6-hourly cadence for consistency.
This model set spans distinct spatial resolutions, training objectives, physical parameterizations, and data assimilation strategies, and therefore exhibits markedly different systematic error characteristics and biases.


\textbf{Baselines and comparisons.}
For each upstream model, we compare (i) the raw deterministic forecast and
(ii) the diffusion-downscaled forecast, and compute \textit{skill scores of the downscaled model relative to the corresponding raw upstream forecast},
\[
\mathrm{RMSESS}
=
1 - \frac{\mathrm{RMSE}_{\text{downscaled}}}{\mathrm{RMSE}_{\text{base}}},
\qquad
\mathrm{CRPSS}
=
1 - \frac{\mathrm{CRPS}_{\text{downscaled}}}{\mathrm{CRPS}_{\text{base}}},
\]
where positive values indicate improvement over the baseline forecast.

\begin{figure}[t]
  \centering
  \begin{subfigure}[t]{0.495\textwidth}
    \centering
    \includegraphics[width=\linewidth]{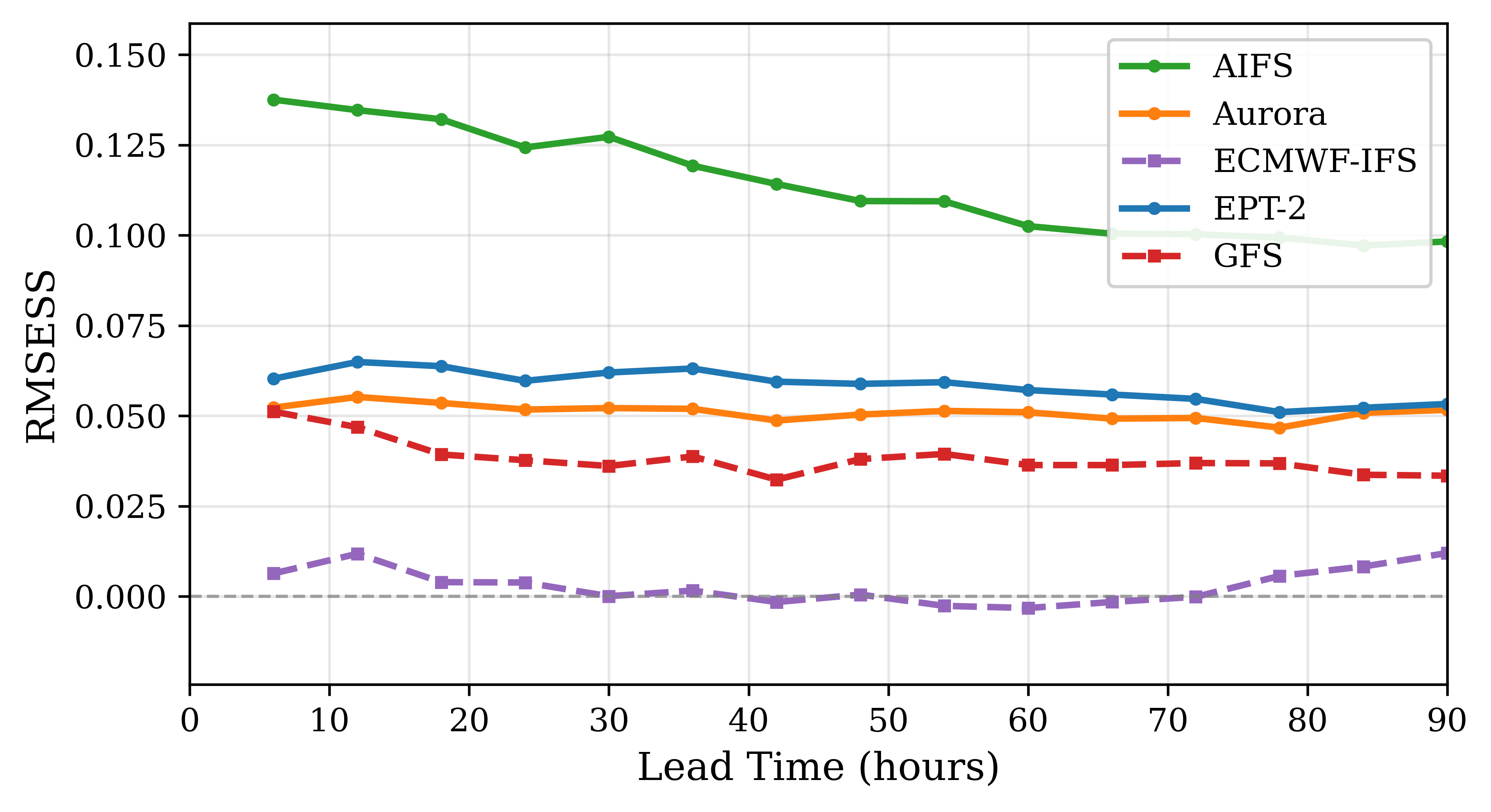}
    \caption{2\,m Temperature}
    \label{fig:rmsess_temp}
  \end{subfigure}
  \hfill
  \begin{subfigure}[t]{0.495\textwidth}
    \centering
    \includegraphics[width=\linewidth]{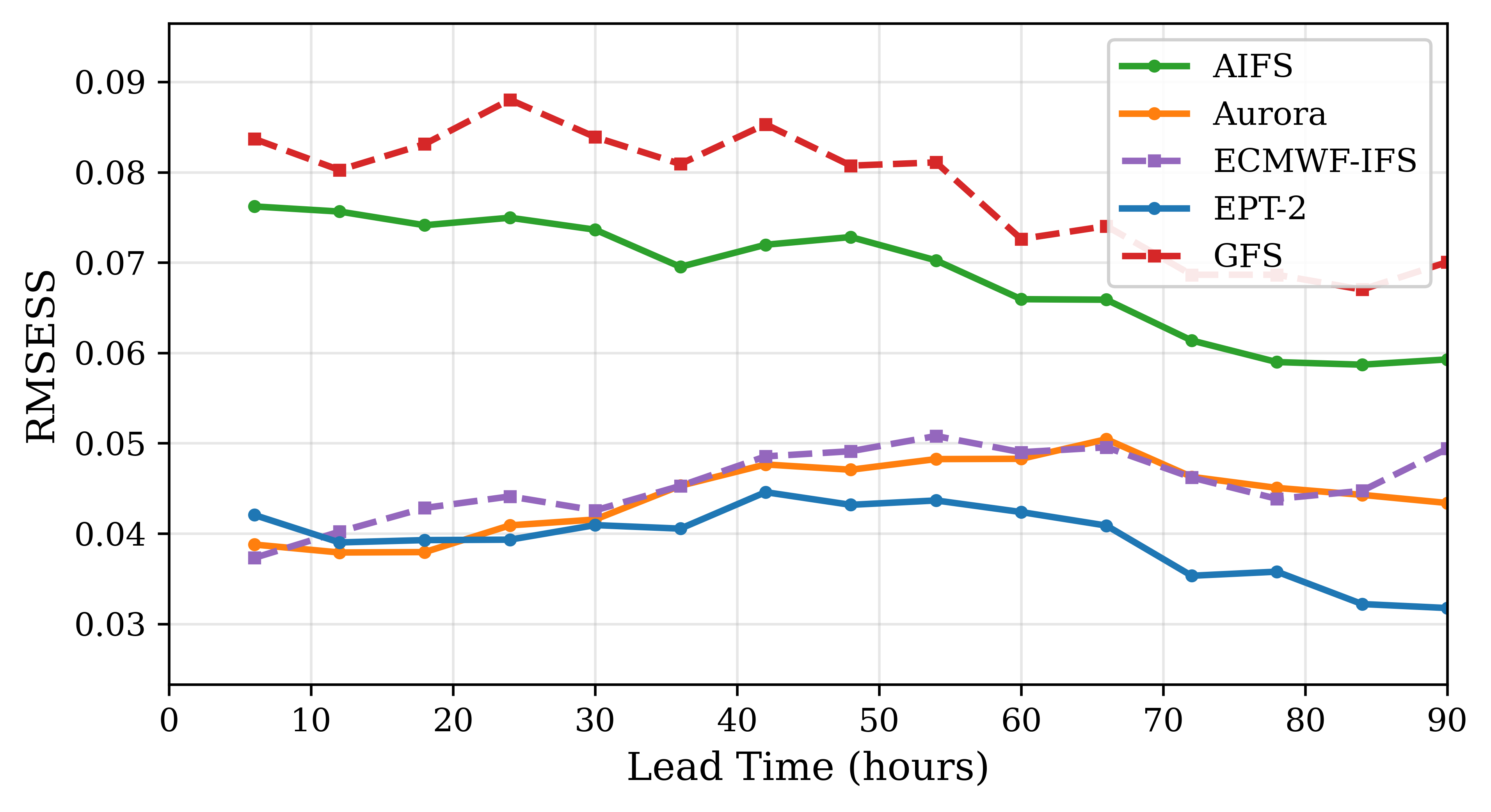}
    \caption{10\,m Wind Speed}
    \label{fig:rmsess_wind}
  \end{subfigure}
  \caption{\textbf{RMSE Skill Score vs.\ lead time.}
  RMSESS ($1 - \text{RMSE}_{\text{downscaled}}/\text{RMSE}_{\text{base}}$) for all models
  (AI: Aurora, AIFS, EPT-2; NWP: GFS, ECMWF IFS) over July~2024--June~2025, up to 90\,h lead time.
  Positive values indicate improvement over the raw baseline.
  The diffusion-based downscaler consistently improves point forecast skill across all
  upstream models.}
  \label{fig:rmsess}
\end{figure}

\textbf{Results.}
Figure~\ref{fig:rmsess} presents the root-mean-square error skill scores (RMSESS)
of the diffusion-based downscaler relative to each upstream model's raw forecast.
Across all evaluated systems, including both AI-based and NWP models, the
downscaler yields predominantly positive RMSESS at nearly all lead times,
indicating systematic improvements in point forecast accuracy.
Although the magnitude of improvement varies by variable, model, and forecast
lead time, positive skill is observed consistently from short-range to
medium-range forecasts and, in most cases, extends beyond day~2.

For 2\,m temperature (left panel of Figure~\ref{fig:rmsess}), AI-based models exhibit the largest relative improvements.
Temperature RMSESS typically ranges from approximately 5--14\% at short lead times and gradually decreases to about 2--6\% at longer lead times. For NWP models, improvements are generally smaller but remain positive, with the
exception of ECMWF-IFS, for which RMSESS approaches zero beyond approximately
24~hours, indicating little to no systematic improvement at longer lead times.

For 10\,m wind speed (right panel of Figure~\ref{fig:rmsess}), relative improvements
are larger and more persistent across lead times.
AI-based models achieve wind RMSESS values of approximately 3--8\% over the full
0--90\,h forecast range.
For NWP models, wind RMSESS remains consistently positive, typically ranging from
about 4--9\% over the 0--90\,h lead times considered.
Notably, RMSE improvements remain positive and not negligible even for strong operational NWP
baselines such as ECMWF IFS  which is currently the world's leading NWP model, demonstrating that the method adds value across
a broad range of upstream forecast qualities.

\begin{figure}[t]
  \centering
  \begin{subfigure}[t]{0.495\textwidth}
    \centering
    \includegraphics[width=\linewidth]{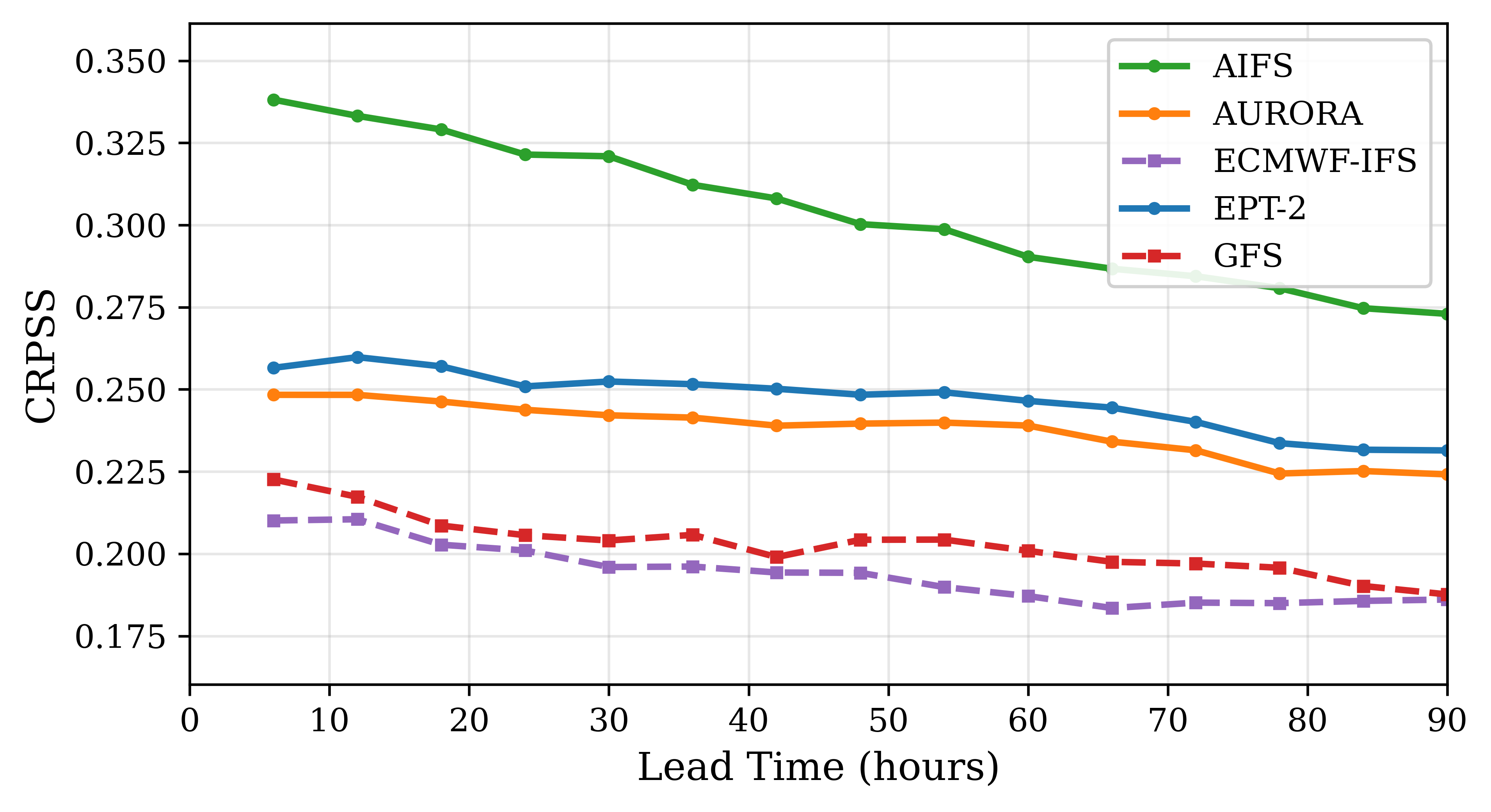}
    \caption{2\,m Temperature}
    \label{fig:crpss_temp}
  \end{subfigure}
  \hfill
  \begin{subfigure}[t]{0.495\textwidth}
    \centering
    \includegraphics[width=\linewidth]{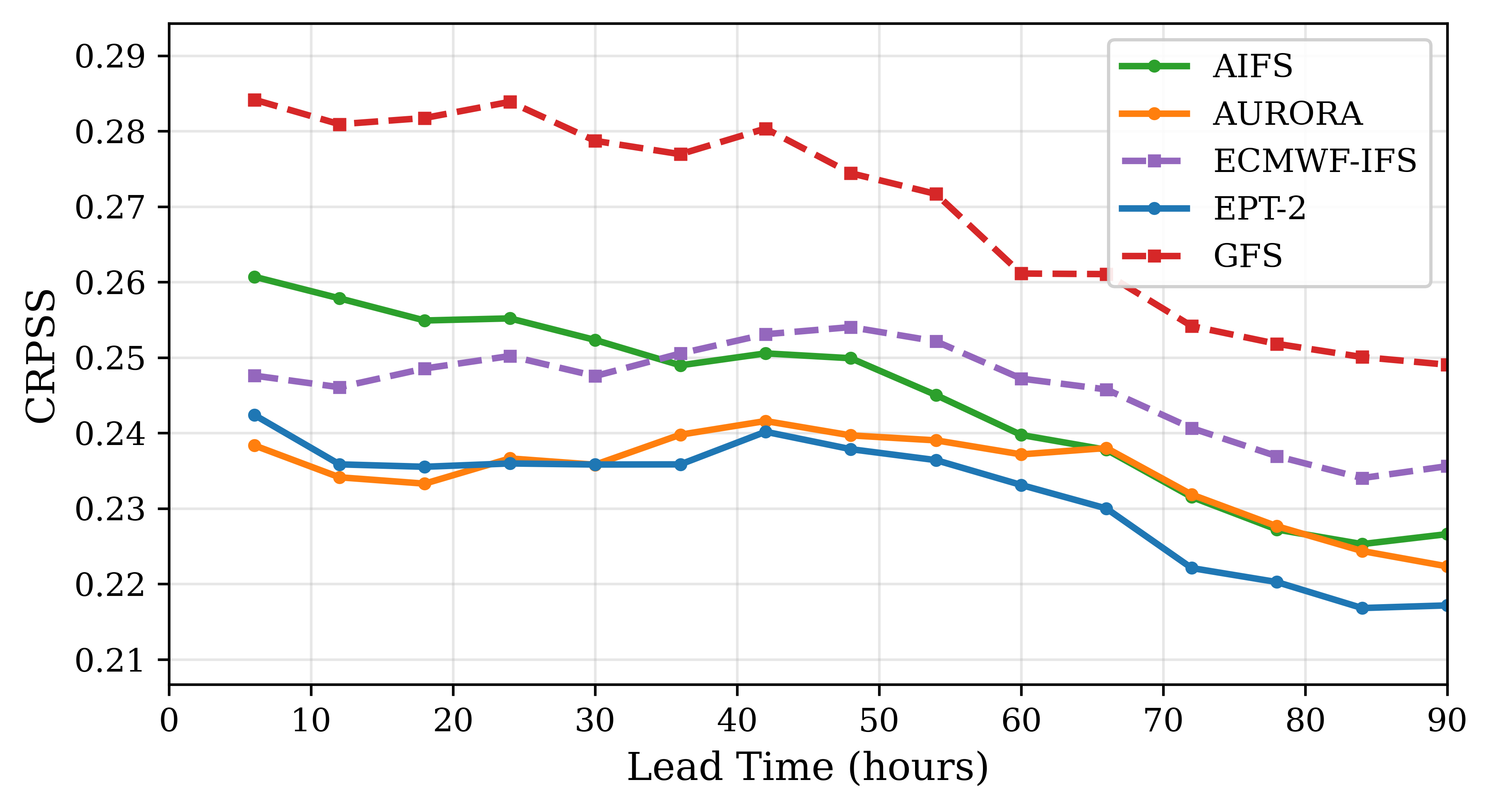}
    \caption{10\,m Wind Speed}
    \label{fig:crpss_wind}
  \end{subfigure}
  \caption{\textbf{CRPS Skill Score vs.\ lead time.}
  CRPSS ($1 - \text{CRPS}_{\text{downscaled}}/\text{CRPS}_{\text{base}}$) for all models
  (AI: Aurora, AIFS, EPT-2; NWP: GFS, ECMWF IFS) over July~2024--June~2025, up to 90\,h lead time.
  Positive values indicate improvement over the deterministic baseline.
  The diffusion-based downscaler improves probabilistic skill across all upstream models.}
  \label{fig:crpss}
\end{figure}

On the probabilistic side, Figure~\ref{fig:crpss} reports continuous ranked
probability skill scores (CRPSS), evaluating the quality of the full predictive
distribution.
Across all upstream models and lead times, diffusion-based downscaling yields
consistently positive CRPSS.
Moreover, the magnitude of probabilistic improvement is substantially larger
than that observed for RMSE.
In the short range, CRPSS frequently reaches values of
15\text{--}30\% for both NWP and AI-based models, with positive
skill persisting at longer lead times.
The systematically larger gains in CRPSS compared to RMSESS indicate that the
primary benefit of diffusion-based downscaling lies in improving the full
conditional distribution rather than only the ensemble mean.

Overall, these results show that diffusion-based downscaling provides a robust,
model-agnostic mechanism for lifting deterministic forecasts into 
high-resolution ensembles, improving both point skill (RMSE of the ensemble mean)
and probabilistic skill (CRPS) in a strictly zero-shot setting across diverse
upstream models, including a strong operational high-resolution baseline.
Sample visualizations comparing CERRA reference fields against ensemble members generated by the diffusion downscaler are provided in Appendix~\ref{sec:appendix:visualizations}.
\paragraph{Training and sampling cost.}
The diffusion downscaler is computationally lightweight both to train and to
deploy.
The model is trained for 50 epochs on eight NVIDIA H100 GPUs, requiring
approximately 8 hours in total.
At inference time, generating 16 high-resolution samples at 5\,km resolution
over the full European domain from a single deterministic forecast step takes
approximately 20 seconds on a single NVIDIA H100 GPU.
Additionally, multiple diffusion
samples can be generated fully in parallel across GPUs.
As a result, probabilistic forecasts can be produced with only a modest increase
in latency relative to the underlying deterministic model.

\begin{figure}[t]
  \centering
  \includegraphics[width=\linewidth]{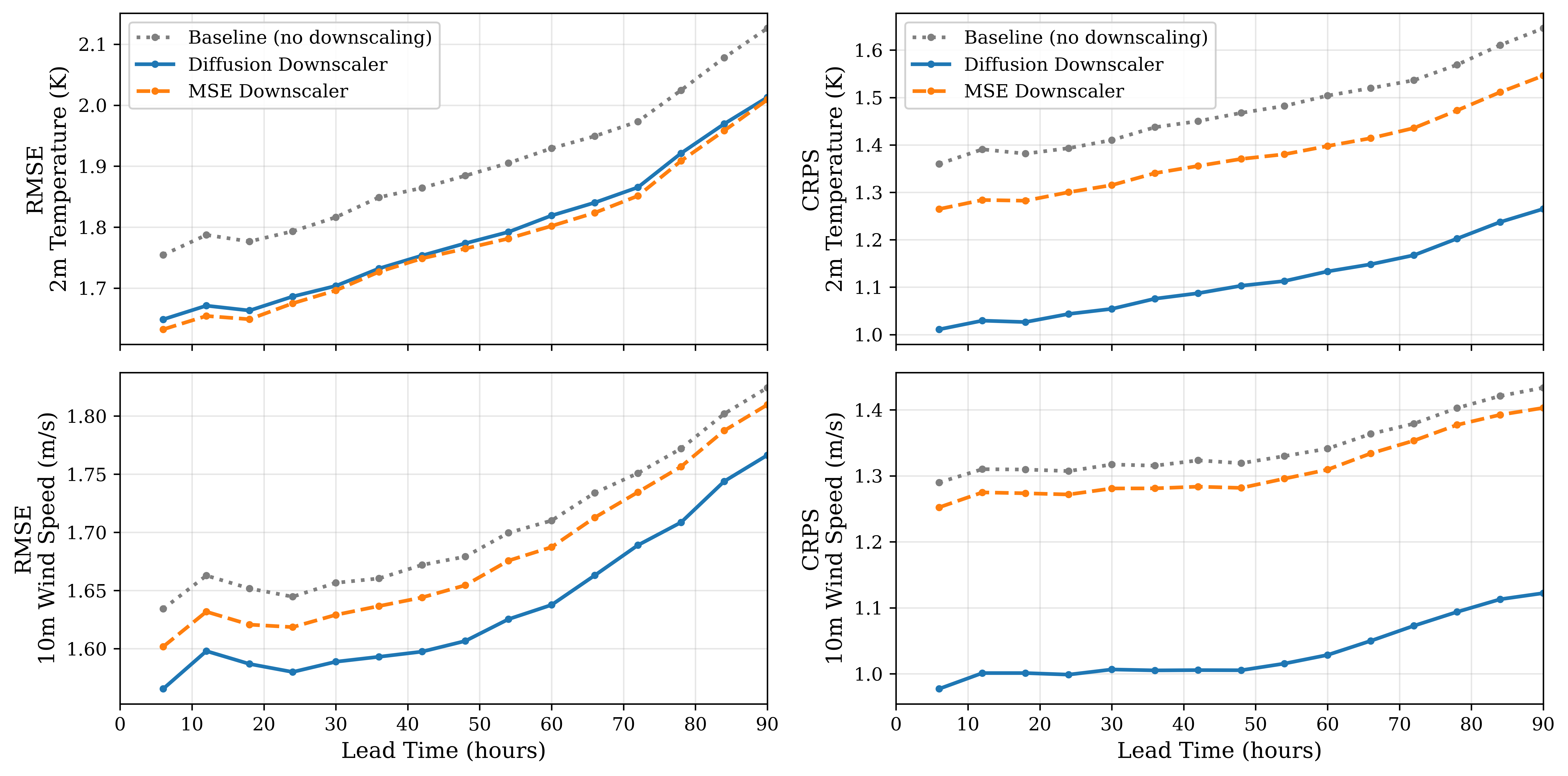}
  \caption{\textbf{Ablation: Diffusion vs.\ regression-based downscaling.}
  RMSE (left) and CRPS (right) for 2\,m temperature (top) and 10\,m wind speed (bottom)
  using EPT-2 as upstream model.
  Dashed: raw EPT-2 forecast.
  Dash-dot: regression-based (MSE) downscaler.
  Solid: diffusion-based downscaler.
  The diffusion model achieves comparable or better RMSE while substantially improving CRPS,
  demonstrating that gains arise from learning a conditional distribution rather than from
  architectural capacity alone.}
  \label{fig:ablation}
\end{figure}

\textbf{Ablation: diffusion vs.\ regression-based downscaling.}
Figure~\ref{fig:ablation} isolates the effect of the diffusion training objective
by comparing against a regression-based downscaler trained with an MSE loss,
using the exact same U-Net architecture.
Both learned downscalers improve RMSE relative to the raw upstream forecast for
2\,m temperature and 10\,m wind speed.
For 10\,m wind speed, the diffusion-based model achieves consistently lower RMSE
than the MSE-based alternative across lead times, whereas for 2\,m temperature
the two approaches exhibit very similar RMSE.
In contrast, pronounced differences emerge in probabilistic skill:
the regression-based model yields only limited CRPS improvements, while the
diffusion-based downscaler achieves substantially lower CRPS across all lead
times and both variables.
This demonstrates that the dominant gains in probabilistic performance stem from
learning a conditional distribution via diffusion, rather than from
architectural capacity alone.

%

\section{Conclusion}
We presented a diffusion-based formulation for probabilistic downscaling that
acts as a \emph{model-agnostic} resolution-lifting operator for deterministic
weather forecasts.
A single conditional EDM-style model is trained once on paired reanalysis fields
(ERA5$\rightarrow$CERRA over Europe) and subsequently applied \emph{zero-shot} to
heterogeneous upstream forecasting systems, spanning both AI-based and numerical
weather prediction (NWP) models.
In all cases, the same trained model is used without any upstream-specific
fine-tuning, producing high-resolution ensembles on the CERRA grid.

In a station-verified, out-of-sample evaluation covering July~2024--June~2025 and
lead times up to 90\,h, diffusion-based downscaling consistently improves
forecast skill relative to each model's own raw deterministic forecast for
near-surface variables, including 2\,m temperature and 10\,m wind speed.
These improvements are observed across upstream models with different numerical
formulations, resolutions, and error characteristics, demonstrating that the
learned downscaler generalizes robustly when used as a post-processing step.

While diffusion-based downscaling yields consistent improvements in point accuracy, 
the magnitude of these gains is modest compared to the substantial and robust 
improvements in probabilistic skill. This reflects the inherently ill-posed nature 
of statistical downscaling, where multiple fine-scale realizations are compatible 
with the same coarse atmospheric state. The primary benefit of the proposed approach 
lies not in sharper deterministic predictions, but in its ability to represent 
flow-dependent uncertainty and generate realistic high-resolution ensembles. 
By operating zero-shot across heterogeneous upstream forecasts, the method provides 
a model-agnostic and computationally efficient pathway for uncertainty-aware regional 
forecasting, producing probabilistic high-resolution fields without the need for 
costly regional dynamical simulations.

\paragraph{Limitations and future work.}
The present study focuses on a limited set of near-surface variables and on a single high-resolution training domain defined by the CERRA European reanalysis.
While this setting is sufficient to demonstrate the feasibility and model-agnostic nature of diffusion-based downscaling, it does not fully probe the method's behavior under stronger forms of distribution shift, such as transfer to climatologically distinct regions.

From an operational perspective, we do not explicitly optimize ensemble size or inference cost, nor do we perform post hoc calibration of ensemble spread. While the reported CRPS improvements indicate enhanced probabilistic skill,
future work will investigate explicit calibration and sharpness diagnostics and their dependence on sampling strategy.

Finally, although the proposed downscaler is intentionally model-agnostic, mild forms of adaptation to upstream forecast structure-such as weak physical constraints, lead-time-dependent conditioning, lightweight per-model
normalization, or limited specialization for regional forecasting systems (e.g., HRRR, ICON-EU, ICON-D2)-may further improve robustness under severe distribution
shift. Exploring such extensions while preserving the core objective of a single,
reusable probabilistic module remains an important direction for future research.

\section*{Acknowledgements}
We gratefully acknowledge the European Centre for Medium-Range Weather Forecasts (ECMWF) and the National Oceanic and Atmospheric Administration (NOAA) for making high-quality reanalysis and forecast data publicly available, without which this work would not have been possible.

\clearpage   

\bibliography{references}
\ifuseiclr
  \bibliographystyle{iclr2026_conference}
\else
  \bibliographystyle{plainnat}
\fi

\appendix
\section{Supplementary Information}
\label{sec:appendix}

\subsection{Spectral smoothing of the conditioning input}
\label{sec:appendix:smoothing}

In this section, we further elaborate on the ERA5 input spectral smoothing mechanism introduced in the main text.

Specifically, the conditioning input $\bar{u}$ is obtained by Fourier-domain
smoothing of the original ERA5 fields, with the smoothing strength sampled
uniformly from the interval $[0,\,0.8]$, where zero corresponds to no smoothing.
This procedure exposes the denoiser to a continuum of effective input
resolutions during training and improves robustness to heterogeneous upstream
models at inference time.

Let $\mathcal{F}$ and $\mathcal{F}^{-1}$ denote the two-dimensional discrete
Fourier transform and its inverse.
We define
\begin{equation}
\label{eq:fourier_smoothing}
\bar{u}
=
\mathcal{F}^{-1}
\left[
\mathcal{F}(\bar{u}_0)
\;\cdot\;
G_{\alpha}
\right],
\end{equation}
where $\cdot$ denotes element-wise multiplication and
$G_{\alpha}$ is an isotropic Gaussian low-pass filter defined in frequency space.

Specifically, letting $(k_x, k_y)$ denote discrete spatial frequencies centered
at zero, the filter takes the form
\begin{equation}
\label{eq:gaussian_filter}
G_{\alpha}(k_x, k_y)
=
\exp\!\left(
-\frac{k_x^2 + k_y^2}{2 \sigma_\alpha^2}
\right),
\end{equation}
with bandwidth
\begin{equation}
\label{eq:sigma_alpha}
\sigma_\alpha
=
\frac{1}{2}
\max\!\left(
1,\,
\frac{\min(H, W)}{2}\,(1 - \alpha)
\right).
\end{equation}

The smoothing strength $\alpha \in [0,\,0.8]$ is sampled uniformly for each
training instance.
The limiting case $\alpha = 0$ corresponds to the identity operator
(no smoothing), while increasing values of $\alpha$ progressively suppress
high-frequency components, yielding conditioning inputs with coarser effective
resolution.

\subsection{Denoiser Training}
\label{sec:appendix:denoiser_training}

The parameters $\theta$ of the conditional denoiser $D_\theta$ are learned from paired low- and high-resolution
reanalysis data.
Training samples are constructed from time-aligned fields, where
$\bar{u}(t)$ denotes a coarse-resolution atmospheric state from ERA5 and
$u(t)$ the corresponding high-resolution target from CERRA at the same valid
time $t$.
Paired samples $(\bar{u},u)$ are obtained by matching timestamps and sampled
from long reanalysis time series; we denote by $\mathcal{D}$ the resulting
empirical distribution of paired fields.

For a given noise level $\sigma>0$ and noise realization
$\eta \sim \mathcal{N}(0,\sigma^2 I)$, a noisy high-resolution field
$u+\eta$ is constructed.
The conditional denoiser $D_\theta(\cdot,\bar{u},\sigma)$ is then trained to
reconstruct the clean target $u$ by minimizing a denoising objective.
Specifically, the denoiser parameters are obtained as (local) minimizers of
\begin{equation}
    \label{eq:dnostr_downscaling}
    \mathcal{L}(D_\theta,\sigma)
    =
    \mathbb{E}_{(\bar{u},u)\sim \mathcal{D}}
    \mathbb{E}_{\eta \sim \mathcal{N}(0,\sigma^2 I)}
    \bigl\|
        D_\theta(u+\eta,\bar{u},\sigma) - u
    \bigr\|_2^2 .
\end{equation}

The full training objective averages this loss over a distribution of noise
levels,
\begin{equation}
    \label{eq:loss_weighted}
    \mathcal{L}(D_\theta)
    =
    \mathbb{E}_{\sigma \sim p_{\mathrm{train}}}
    \left[
        \lambda(\sigma)\,\mathcal{L}(D_\theta,\sigma)
    \right],
\end{equation}
where $p_{\mathrm{train}}$ denotes the noise-level sampling distribution and
$\lambda(\sigma)$ a noise-dependent weighting factor.

\paragraph{Preconditioning and effective weighting.}
\label{sec:appendix:precond}
For all experiments, we adopt the preconditioning and weighting scheme of
Elucidated Diffusion Models (EDM).
The denoiser is parameterized as
\begin{equation}
    \label{eq:precond_denoiser}
    D_\theta(u+\eta,\bar{u},\sigma)
    =
    c_{\mathrm{skip}}(\sigma)(u+\eta)
    +
    c_{\mathrm{out}}(\sigma)\,
    F_\theta\!\left(
        c_{\mathrm{in}}(\sigma)(u+\eta),\,
        \bar{u};\,
        c_{\mathrm{noise}}(\sigma)
    \right),
\end{equation}
where $F_\theta$ denotes the raw neural network.

The preconditioning coefficients are given by
\begin{align}
    c_{\mathrm{skip}}(\sigma) &= \frac{\sigma_{\mathrm{data}}^2}{\sigma^2 + \sigma_{\mathrm{data}}^2}, \\
    c_{\mathrm{out}}(\sigma)  &= \frac{\sigma\,\sigma_{\mathrm{data}}}{\sqrt{\sigma^2 + \sigma_{\mathrm{data}}^2}}, \\
    c_{\mathrm{in}}(\sigma)   &= \frac{1}{\sqrt{\sigma^2 + \sigma_{\mathrm{data}}^2}}, \\
    c_{\mathrm{noise}}(\sigma) &= \tfrac{1}{4}\log(\sigma),
\end{align}
with $\sigma_{\mathrm{data}} = 0.5$.

The corresponding regression target is
\begin{equation}
    F_{\mathrm{target}}
    =
    \frac{1}{c_{\mathrm{out}}(\sigma)}
    \left(
        u - c_{\mathrm{skip}}(\sigma)(u+\eta)
    \right),
\end{equation}
and the resulting training loss can be written as
\begin{equation}
    \mathcal{L}
    =
    \mathbb{E}_{\sigma,(\bar{u},u),\eta}
    \left[
        w(\sigma)
        \left\|
            F_\theta\!\left(
                c_{\mathrm{in}}(\sigma)(u+\eta),\,
                \bar{u};\,
                c_{\mathrm{noise}}(\sigma)
            \right)
            -
            F_{\mathrm{target}}
        \right\|_2^2
    \right],
\end{equation}
with effective weighting
\begin{equation}
    w(\sigma)
    =
    \frac{\sigma^2 + \sigma_{\mathrm{data}}^2}{(\sigma\,\sigma_{\mathrm{data}})^2}.
\end{equation}

Noise levels are sampled during training according to a log-normal
distribution,
\begin{equation}
    \log \sigma \sim \mathcal{N}(P_{\mathrm{mean}}, P_{\mathrm{std}}^2),
\end{equation}
with $P_{\mathrm{mean}}=-0.5$ and $P_{\mathrm{std}}=1.5$ in all experiments.

\subsubsection{Inference and sample generation}
\label{sec:appendix:inference}

At inference time, samples from the learned conditional distribution
$p_\theta(u \mid \bar{u})$ are generated by reversing the diffusion process.
Starting from the conditional reverse-time stochastic differential equation Eq.(\ref{eq:reverse_sde_downscaling}), we consider its associated \textit{deterministic} probability-flow formulation:

\begin{equation}
\label{eq:pf_ode_downscaling}
\mathrm{d}x
=
\left[
    \frac{\dot{\sigma}(\tau)}{\sigma(\tau)}
    +
    \frac{\dot{s}(\tau)}{s(\tau)}
\right] x\,\mathrm{d}\tau
-
\frac{s(\tau)\,\dot{\sigma}(\tau)}{\sigma(\tau)}
D_\theta\!\left(
    \frac{x}{s(\tau)},\,
    \bar{u},\,
    \sigma(\tau)
\right)\mathrm{d}\tau,
\end{equation}
where the score function is approximated by the trained conditional denoiser
$D_\theta$.
Both the stochastic formulation in Eq.~\eqref{eq:reverse_sde_downscaling} and the
deterministic probability-flow ODE in Eq.~\eqref{eq:pf_ode_downscaling} induce the
same family of conditional distributions $p_\tau(u \mid \bar{u})$.

Following the Elucidated Diffusion Models (EDM) framework \citep{karras2022elucidating}, we adopt a
variance-exploding diffusion with identity signal scaling,
\begin{equation}
    s(\tau) = 1,
    \qquad
    \sigma(\tau) = \tau,
\end{equation}
so that diffusion time directly parameterizes the noise level and
Eq.~\eqref{eq:pf_ode_downscaling} simplifies accordingly.

In practice, inference is performed by discretizing the probability-flow ODE over a
monotonically decreasing sequence of noise levels
$\{\sigma_i\}_{i=1}^N \subset [\sigma_{\max}, \sigma_{\min}]$.
The discrete noise levels are constructed using the EDM power-law schedule,
\begin{equation}
    \sigma_i
    =
    \left(
        \sigma_{\max}^{1/\rho}
        +
        \frac{i-1}{N-1}
        \left(
            \sigma_{\min}^{1/\rho}
            -
            \sigma_{\max}^{1/\rho}
        \right)
    \right)^{\rho},
    \qquad i = 1,\dots,N,
\end{equation}
with $N = 128$, $\sigma_{\max} = 80$, and $\sigma_{\min} = 0.002$.

The resulting ODE is integrated using a first-order explicit Euler method. Integration is initialized from
\begin{equation}
    x_{\sigma_{\max}} \sim \mathcal{N}(0,\sigma_{\max}^2 I),
\end{equation}
and proceeds until $\sigma_{\min}$, at which point the final state is taken as a
sample from the conditional distribution $p_\theta(u \mid \bar{u})$.
Multiple independent realizations are obtained by repeating the integration with
different initial noise samples.

\subsection{Denoiser Architecture and Configuration}
\label{sec:denoiser_arch}

The denoiser $D_\theta$ is implemented as a 2D encoder--decoder U-Net operating
on latitude--longitude fields on the CERRA grid (Figure~\ref{fig:denoiser_schematic}).
At each forecast lead time, the network takes as input the concatenation
\(
x_\tau = \mathrm{Concat}[u_\tau,\bar{u},s]
\),
where $u_\tau\in\R^{H\times W\times 4}$ is the noisy high-resolution sample
(t2m, u10, v10, msl), $\bar{u}\in\R^{H\times W\times 4}$ is the bilinearly
upsampled coarse forecast, and $s\in\R^{H\times W\times 2}$ contains static
fields (elevation and land--sea mask). Hence $x_\tau\in\R^{H\times W\times 10}$.

\paragraph{Encoder--decoder layout.}
The network begins with an input projection (3$\times$3 convolution) to
$C_0{=}32$ channels.
It then applies $L{=}4$ resolution stages with channel multipliers
\(
[2,4,8,8]
\),
resulting in channel widths
\[
C_0 \rightarrow 2C_0 \rightarrow 4C_0 \rightarrow 8C_0 \rightarrow 8C_0
\quad\text{i.e.}\quad
32 \rightarrow 64 \rightarrow 128 \rightarrow 256 \rightarrow 256.
\]
Downsampling between stages is performed by a strided convolution (kernel 3,
stride 2), halving spatial resolution at each stage.
The decoder mirrors the encoder with symmetric upsampling stages.
Upsampling is performed via bilinear interpolation by a factor of $2$, followed
by a 3$\times$3 convolution.

\paragraph{Residual blocks and skip connections.}
Each resolution stage contains $n_{\mathrm{res}}{=}2$ residual blocks
(Figure~\ref{fig:denoiser_schematic}, blue ``T'' blocks), and standard U-Net skip
connections connect encoder and decoder features at matching resolutions.
Within each residual block, we use GroupNorm--activation--Conv twice (GN--$f$--C,
GN--$f$--C), with dropout set to $0$.
Unless stated otherwise, GroupNorm uses 32 groups.

\paragraph{Conditioning interface.}
Diffusion noise conditioning is injected into every residual block through
adaptive scale--shift modulation.
We embed $\sigma$ using Fourier features with embedding dimension
$d_\sigma{=}256$, followed by a two-layer MLP that outputs per-channel affine
parameters $(a,b)$.
These parameters modulate a normalization layer in FiLM form,
\begin{equation}
    \label{eq:film_gn}
    \mathrm{GN}(x)\;\mapsto\; (1+a)\,\mathrm{GN}(x) + b,
\end{equation}
where $(a,b)$ are broadcast over spatial dimensions.

\paragraph{Attention blocks.}
The denoiser follows a standard U-Net encoder--decoder structure with residual
blocks at each resolution and a single self-attention block at the bottleneck
(see Figure~\ref{fig:denoiser_schematic}).
Self-attention is not used at intermediate encoder or decoder resolutions; all
attention-based global mixing is confined to the bottleneck representation.

\paragraph{Configuration (for reproducibility).}
For clarity, we summarize the key architectural hyperparameters used throughout:
\begin{itemize}
    \item Input channels: $10$ (noisy high-resolution sample $u_\tau$: 4;
    coarse-resolution forecast provided as conditioning: 6).
    \item Output channels: $4$ (2\,m temperature, 10\,m wind components, and mean
    sea-level pressure on the CERRA grid).
    \item Base channels: $C_0 = 32$, with channel multipliers $[2,4,8,8]$,
    corresponding to feature widths
    $32 \rightarrow 64 \rightarrow 128 \rightarrow 256 \rightarrow 256$.
    \item Residual blocks per resolution: $n_{\mathrm{res}} = 2$; dropout $=0$.
    \item Noise embedding dimension: $256$, used to generate adaptive scale and
    shift parameters for conditioning normalization layers.
\end{itemize}
Overall, the model counts a total number of 24M parameters.

\subsection{Denoiser Training Details}
During training, we used paired samples from ERA5 and CERRA at three-hourly resolution, spanning the years 2014–2023, resulting in 26280 samples. We used the AdamW optimizer  \citep{loshchilov_decoupled_2019} with a learning rate of $10^{-4}$, a weight decay of $10^{-5}$, and cosine anneal the learning rate to  $10^{-5}$. The batch size was set to 1.

\subsection{Sample Visualizations}
\label{sec:appendix:visualizations}

\begin{figure}[htbp]
    \centering
    \begin{subfigure}[b]{0.495\textwidth}
        \centering
        \includegraphics[width=\textwidth]{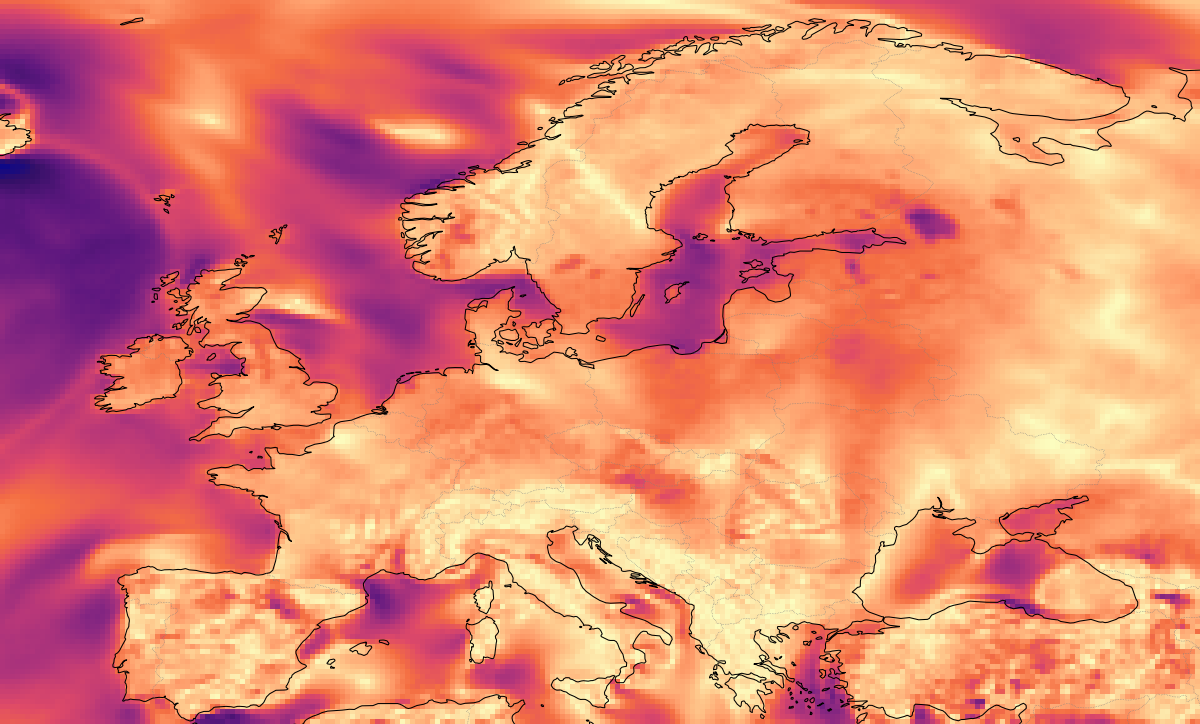}
        \caption{Upstream forecast}
    \end{subfigure}
    \hfill
    \begin{subfigure}[b]{0.495\textwidth}
        \centering
        \includegraphics[width=\textwidth]{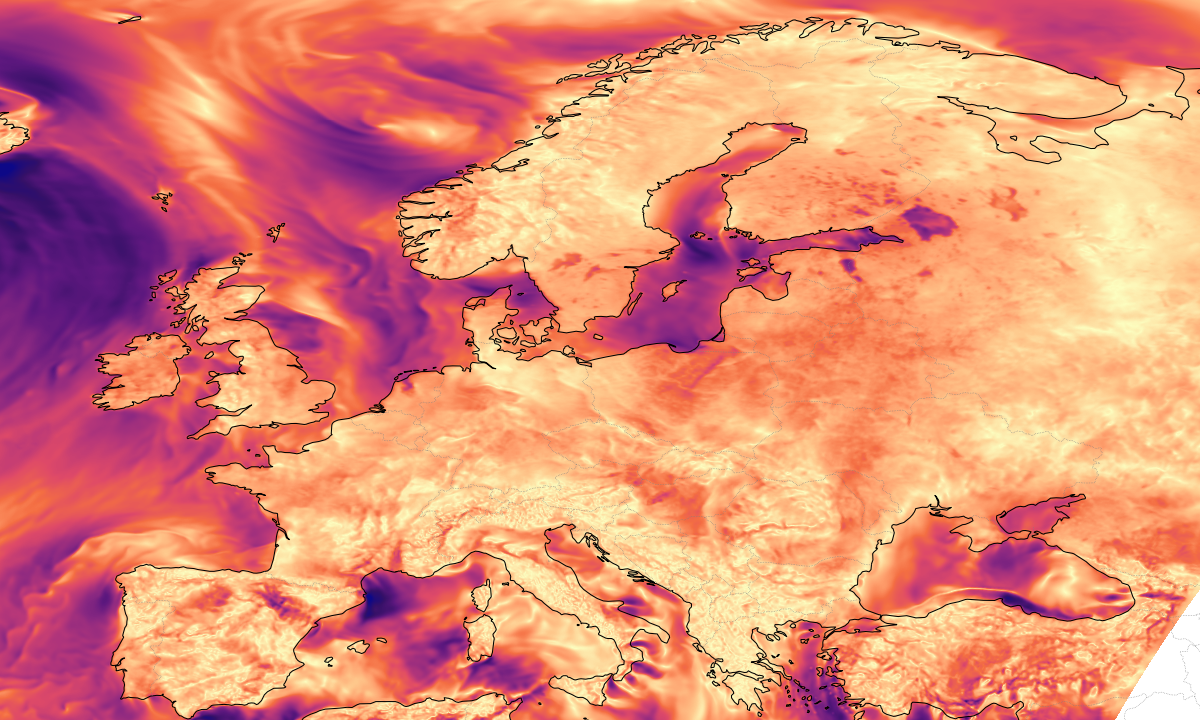}
        \caption{CERRA reference}
    \end{subfigure}
    
    \vspace{0.5em}
    
    \begin{subfigure}[b]{0.495\textwidth}
        \centering
        \includegraphics[width=\textwidth]{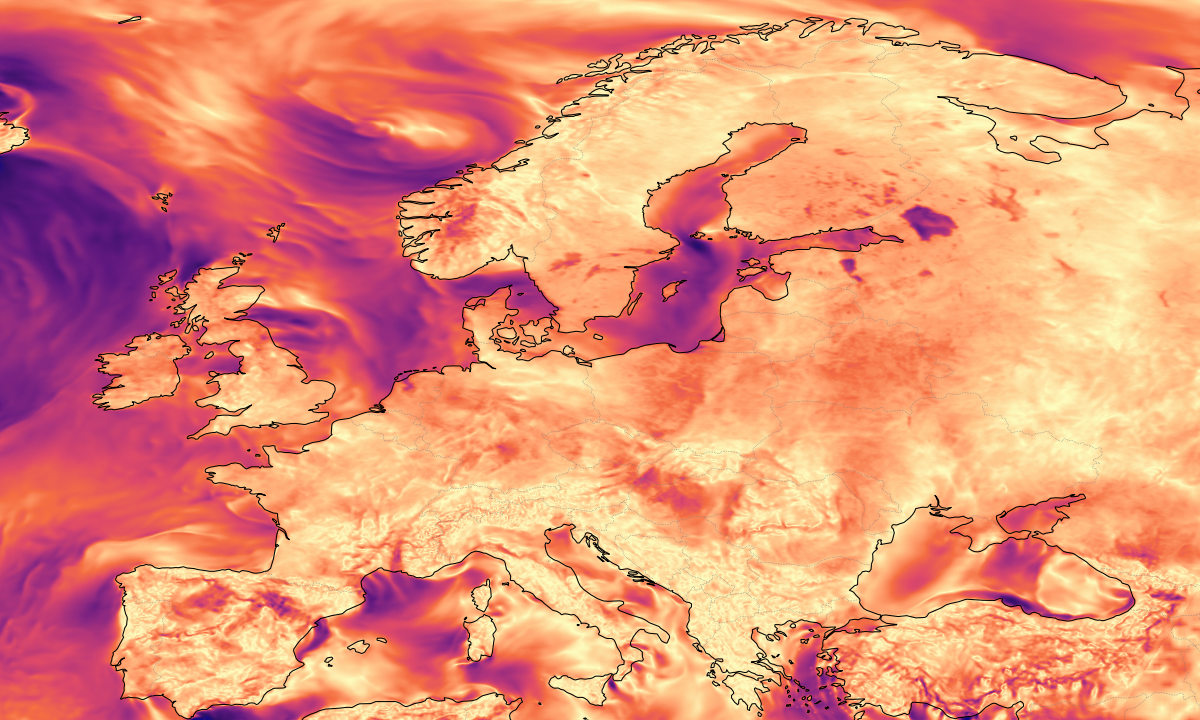}
        \caption{Ensemble member 1}
    \end{subfigure}
    \hfill
    \begin{subfigure}[b]{0.495\textwidth}
        \centering
        \includegraphics[width=\textwidth]{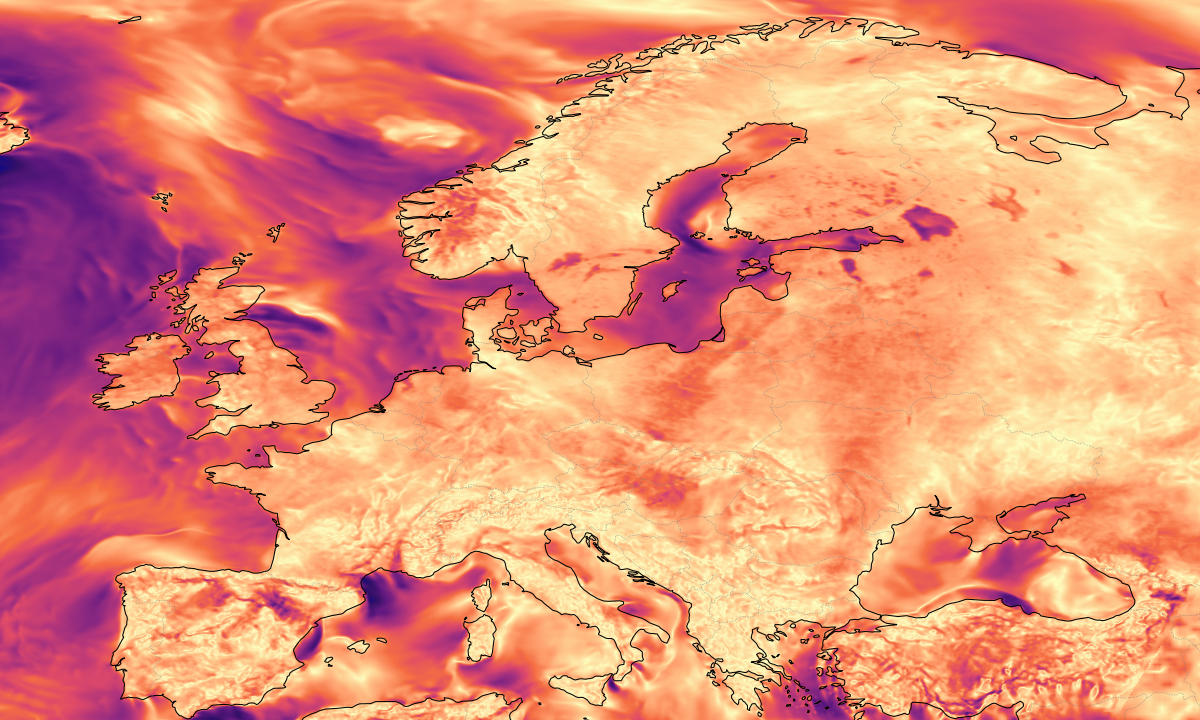}
        \caption{Ensemble member 2}
    \end{subfigure}
    
    \caption{10\,m wind speed for 1 July 2024 at 6-hour lead time. (a) Coarse upstream forecast, (b) CERRA reference, (c--d) two independent ensemble members from the diffusion downscaler.}
    \label{fig:viz_wind}
\end{figure}

\begin{figure}[htbp]
    \centering
    \begin{subfigure}[b]{0.495\textwidth}
        \centering
        \includegraphics[width=\textwidth]{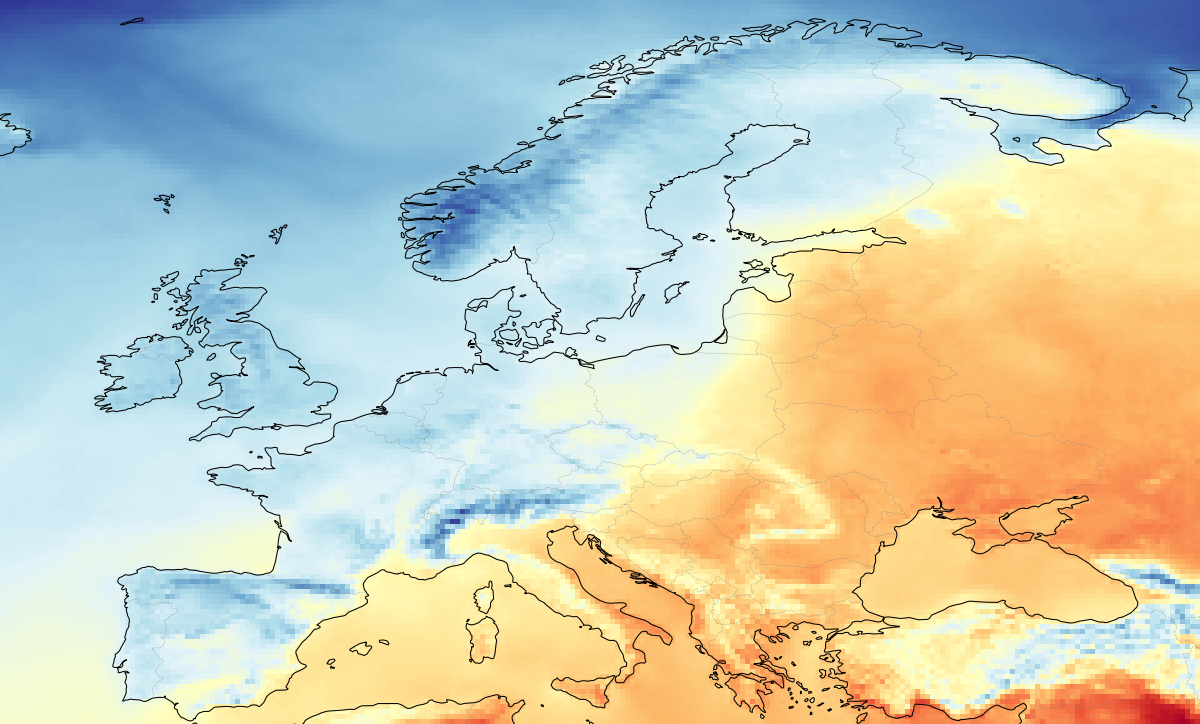}
        \caption{Upstream forecast}
    \end{subfigure}
    \hfill
    \begin{subfigure}[b]{0.495\textwidth}
        \centering
        \includegraphics[width=\textwidth]{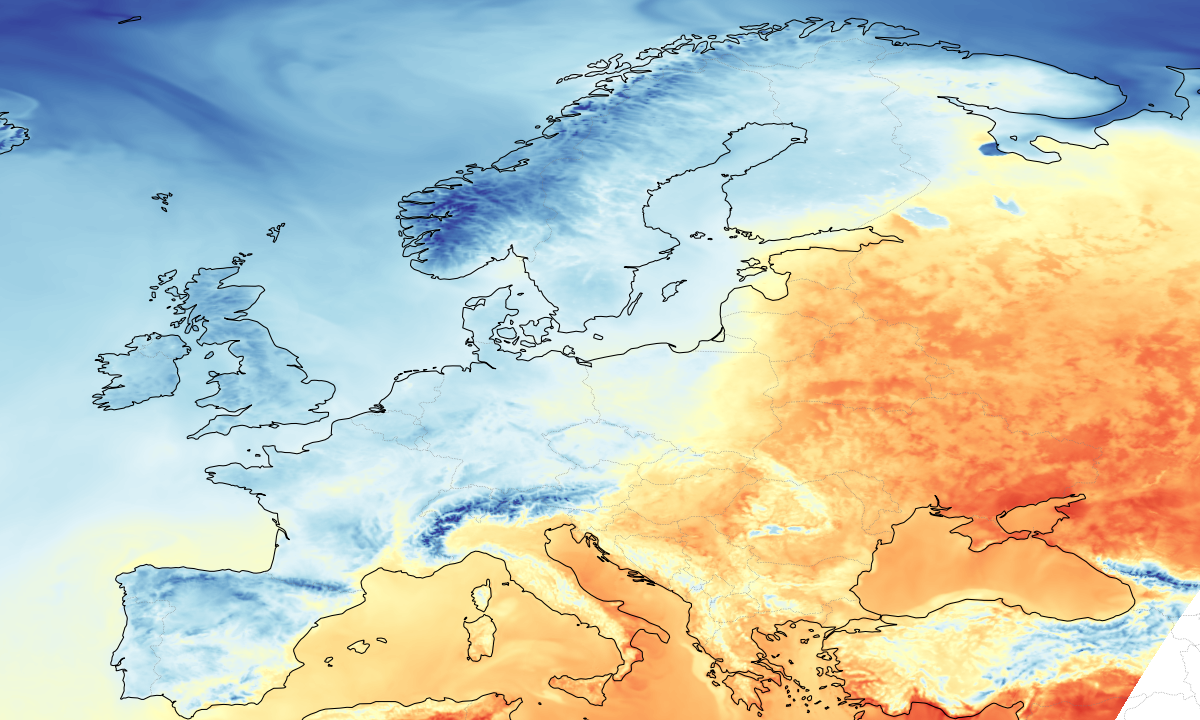}
        \caption{CERRA reference}
    \end{subfigure}
    
    \vspace{0.5em}
    
    \begin{subfigure}[b]{0.495\textwidth}
        \centering
        \includegraphics[width=\textwidth]{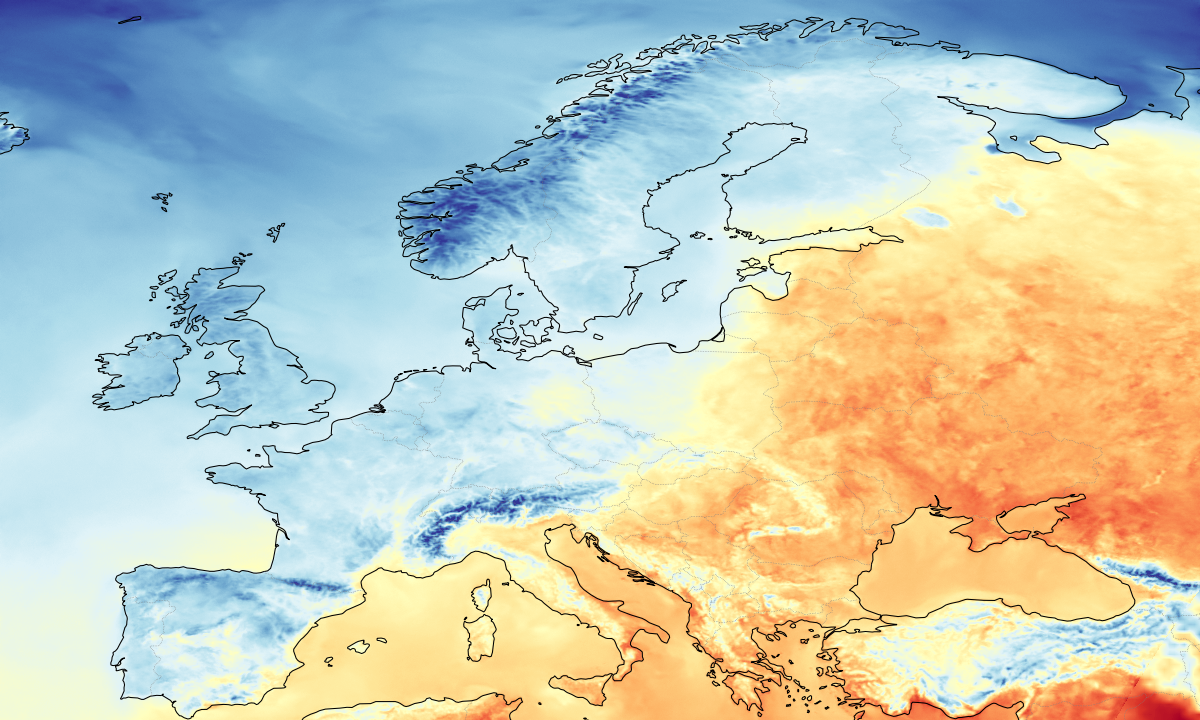}
        \caption{Ensemble member 1}
    \end{subfigure}
    \hfill
    \begin{subfigure}[b]{0.495\textwidth}
        \centering
        \includegraphics[width=\textwidth]{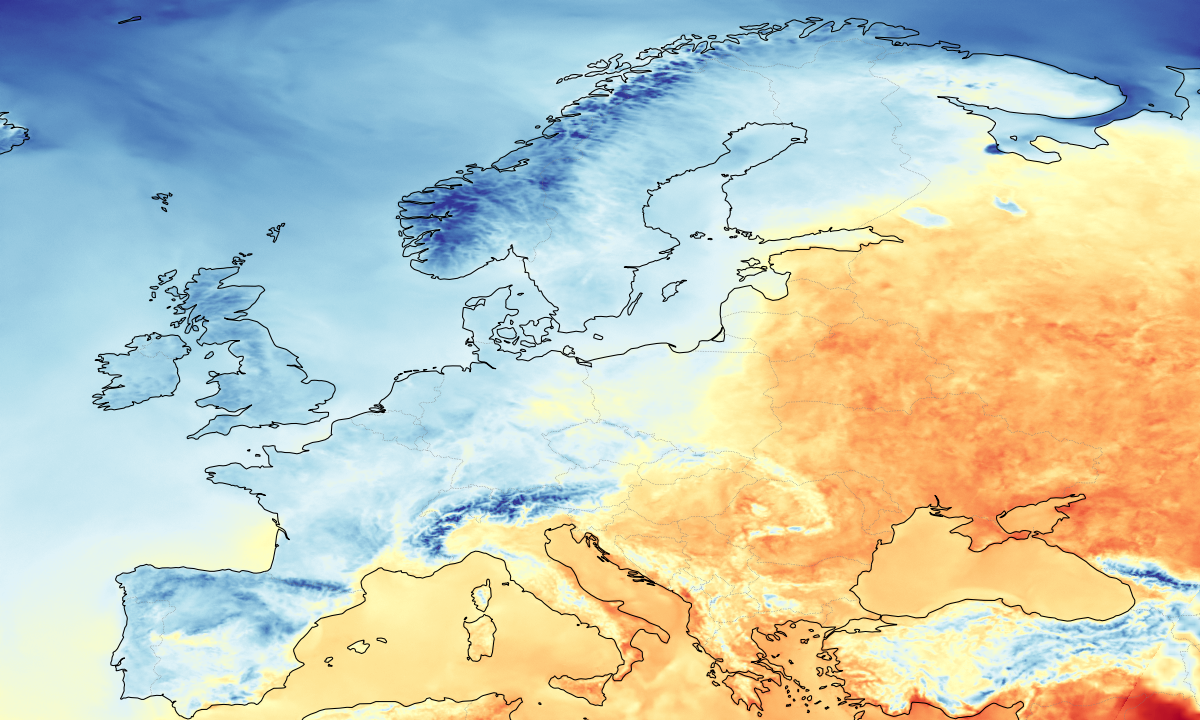}
        \caption{Ensemble member 2}
    \end{subfigure}
    
    \caption{2\,m temperature for 1 July 2024 at 6-hour lead time. (a) Coarse upstream forecast, (b) CERRA reference, (c--d) two independent ensemble members from the diffusion downscaler.}
    \label{fig:viz_temp}
\end{figure}

Figures~\ref{fig:viz_wind} and~\ref{fig:viz_temp} show example downscaled fields for 10\,m wind speed and 2\,m temperature at a 6-hour lead time, comparing the coarse upstream forecast, the CERRA reference, and two independent ensemble members generated by our diffusion model.

\end{document}